\documentclass[11pt]{article}

\usepackage[final]{acl}

\usepackage{times}
\usepackage{latexsym}

\usepackage[T1]{fontenc}

\usepackage[utf8]{inputenc}

\usepackage{microtype}

\usepackage{inconsolata}

\usepackage{graphicx}

\usepackage{amsmath}
\usepackage{amssymb}
\usepackage{makecell}
\usepackage{multirow}
\usepackage{subcaption}

\title{Learning Moral Diversity: Modelling Individual Perspectives in Moral Classification of Texts}

\author{Yi Ren, Lewis Mitchell, Matthew Roughan \\
        School of Mathematical Sciences, Adelaide University\\
        \texttt{\{yi.ren, lewis.mitchell, matthew.roughan\}@adelaide.edu.au}}

\begin{document}
\maketitle

\begin{abstract}

Understanding moral values in social media text offers insight into moral judgement formation, and supervised NLP models trained on crowdsourced data have achieved strong classification performance. However, most approaches simplify the problem by aggregating multiple annotators’ labels into a single ``ground truth”, overlooking the inherent subjectivity of the task. In practice, there are  disagreements between annotators caused by personal viewpoint or inherent ambiguities, particularly for short tweets. Here, we extend a pretrained language model with a layer that learns annotator-specific features. Our model improves predictions of individual annotations and yields representations that reveal meaningful insights into annotators’ moral perspectives. We show that models trained on aggregated labels may hide variation and give a misleading impression of performance. Overall, we demonstrate that disagreement reflects the inherent subjectivity of the task and that modelling individual perspectives creates benefits for moral classification of texts.

\end{abstract}

\section{Introduction}

Morality plays a vital role in shaping people's opinions and forming judgement towards social events. Accordingly, analysing morality allows for better understanding of what people believe and how people interact and form communities. 
We can explore these beliefs through opinions and stances expressed via language, in particular online content. Such analysis has inspired diverse research directions---from analysing political ideology and polarisation \cite{Haidt2007-libvscon} to understanding how people engage in public-health discourse \cite{covid_nlp_network, covid19_vaccine}. Thus, it is extremely valuable to be able to extract moral values from human-created content.

\begin{figure}
    \centering
    \includegraphics[width=\linewidth, trim={6cm 3cm 6cm 3cm}, clip]{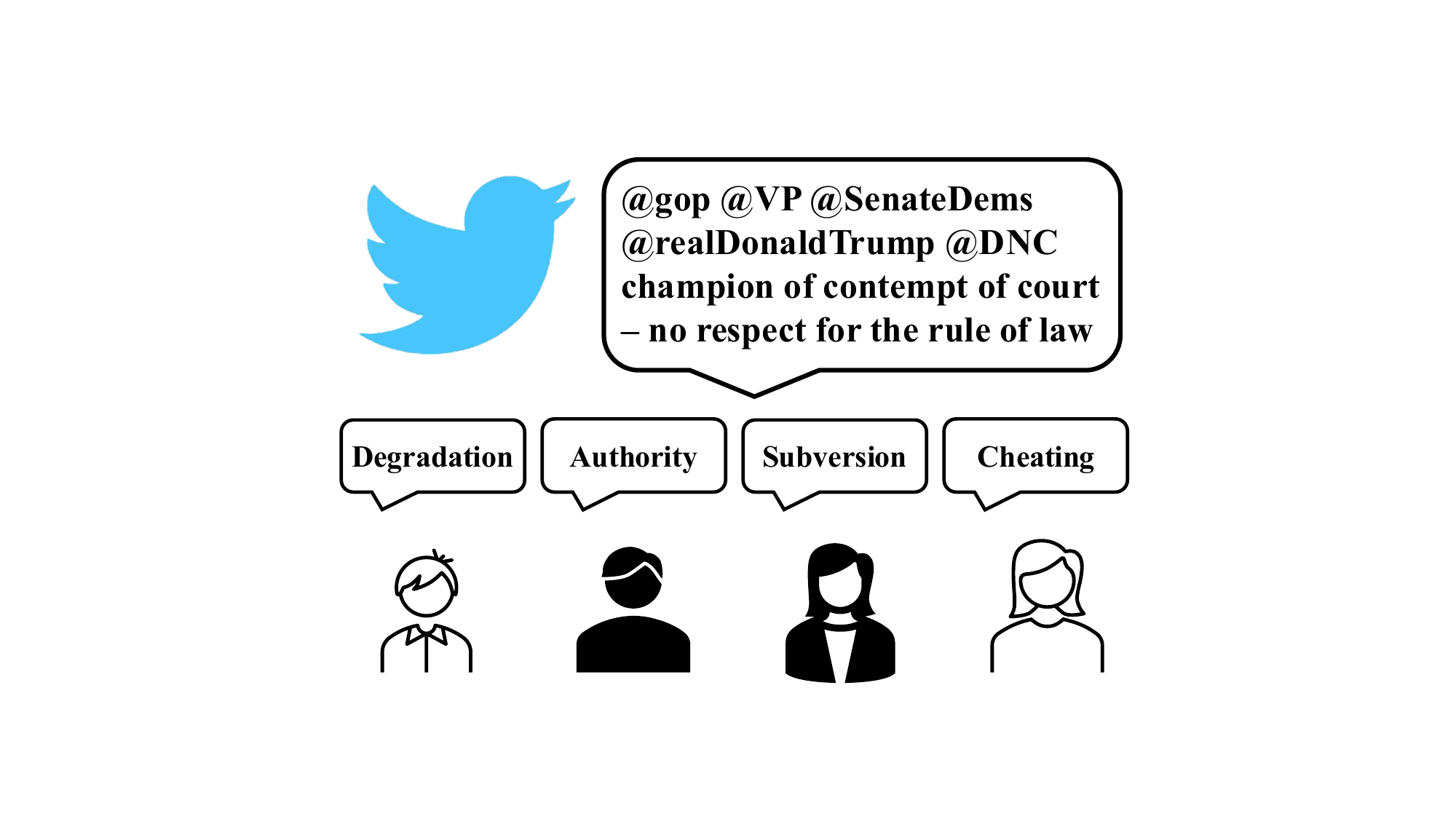}
    \caption{A tweet example with disagreeing labels given by four annotators in the Moral Foundations Twitter Corpus \cite{MFTC}, demonstrating how individuals interpret moral expressions differently.}
    \label{tweet annotation example}
\end{figure}

Many Natural Language Processing (NLP) techniques have been applied and integrated with an empirically validated psychological framework called \textbf{Moral Foundations Theory} (MFT) \cite{Graham2009-da, Graham2013, haidt2012righteous}. MFT decomposes human beliefs into five moral foundations, each with its own virtue and vice axis: \textit{Authority/Subversion}, \textit{Care/Harm}, \textit{Fairness/Cheating}, \textit{Loyalty/Betrayal} and \textit{Purity/Degradation}. Approaches including lexicons \cite{moralstrength, frimermfd2.0, graham2012mfd, eMFD} and supervised machine learning models \cite{Beiro2022-sk, Huang2022-ty, background_acquiring} have been employed to classify text according to MFT.

Recently, transformer-based large language models have shown promising results in classification tasks. BERT in particular have been widely applied due to its ability to generate rich contextual and semantic embeddings \cite{BERToriginal}. Fine-tuning BERT with crowdsourced data has become a standard and effective approach for achieving state-of-the-art performance in moral value classifications \cite{Guo2023-de, Mformer, MoralBERT}. 

However, moral judgement is inherently subjective---individuals hold different beliefs and inhabit different contexts that lead to them interpreting content differently and prioritising different foundations. Additionally, classifying texts from social discourse further extends the subjectivity due to the ambiguity of text data, particularly in highly abbreviated contexts such as tweets. Hence, crowdsourced training datasets in this field often exhibit substantial disagreement among annotators (Figure \ref{tweet annotation example}), yet existing work typically disregards this information, treating disagreement as unstructured noise. They typically train a universal classifier treating moral classification as if there exists a “ground truth” that can be derived from aggregated annotators' responses.

In reality, the disagreements are an important part of the content of the analysis of morality. A model trained to derive one ``smoothed out" response will miss the inherent conflicts and subtleties that are so much of the human experience. A more sophisticated approach is to build models that learn how individuals differ in their judgement.  

In this work, we take a step towards this goal by training classifiers that model annotators in crowdsourced data specifically. We do this by adding a neural network layer on top of finetuning of BERT. This additional layer learns how a particular annotator interprets moral content differently from the shared text embeddings. Furthermore, the added layer can be seen as an interpretable representation that capture meaningful differences between individuals, revealing biases and tendencies towards different moral values. Our results show substantial improvement in predicting individual annotations and highlight concerns that training classifiers on aggregated labels may appear highly accurate but mask inconsistencies across annotators.

This work makes three key contributions:
\begin{enumerate}
    \item A modelling approach that captures individual perspectives in moral value classification, accounting for task subjectivity;
    \item We demonstrate that annotator biases and tendencies in crowdsourced datasets are learnable features rather than noise, where our model yields on average, a 10.2 \% improvement in classification accuracy when compared with finetuned BERT over five foundations; and
    \item  We raise concerns about aggregating annotations into a single ground truth label in such task, urging future modelling approaches to incorporate individual-level variation.
\end{enumerate}

\section{Related Work}

\subsection{NLP for MFT}

\noindent Many studies focus on supervised learning approaches where classifiers are trained to map texts to moral values. Classical machine learning models such as logistic regression and support vector machines, as well as deep learning models such as long short-term memory networks, have been widely adopted \cite{ moralstrength, Beiro2022-sk, MFTC, background_acquiring, MFRC}. More recent work uses transformer-based pretrained language models, particularly BERT and its variants. \citet{MFRC} report baseline performance from finetuning BERT models on a labelled dataset of Reddit posts. \citet{Mformer} and \citet{MoralBERT} provide in-depth analyses of BERT fine-tuning procedures and evaluations of performance in practice. Supervised learning approaches have shown promising results in classification tasks. However, training of supervised models typically rely on large-scale crowdsourced datasets that often exhibit annotation disagreement (as noted earlier). Existing works often handle this disagreement through label aggregation, implicitly assuming the existence of a ground-truth label that can be derived through summarisation or averaging of the various inputs. Yet studies in moral psychology demonstrate that moral judgement is inherently subjective and varies across individuals \cite{haidt2012righteous}, indicating that aggregation may hide meaningful differences in how people interpret the same content.

\subsection{Subjectivity in Moral Judgement}

Human moral judgement is widely recognised as subjective and shaped by individual differences. Haidt’s social intuitionist model highlights how moral judgements arise from intuitive, socially and culturally shaped processes \citep{Haidt2001-lm}. This model later informed MFT, which underpins most NLP research on moral value classification and posits that moral judgement varies across individuals \cite{haidt2012righteous}. Cultural differences were among the most influential factors shaping variability in moral judgement. Early work shows that people from different countries make different moral evaluations, even when presented with the same scenarios \cite{Haidt1993-fl}. Subsequent studies using the Moral Foundations Questionnaires further validated the impact of demographic and cultural differences in moral intuitions \cite{Atari2023-MFQ2, Graham2011-MFQ}. Furthermore, studies also show that interpretation of morality shifts depends on social identities \cite{Ellemers2015-og, Koleva2012-kd}. One must then consider individual perspectives when forming moral judgement. \citet{liscio-etal-2022-cross} explicitly analyse the relationship between model performance and annotator agreement, and suggest modelling approaches to incorporate annotator (dis-)agreement.

\subsection{Data Perspectivism}
  
Recent work on subjective NLP tasks has increasingly challenged the assumption of having a single``ground truth" and the use of aggregated labels, advocating instead for modelling individual annotator perspectives. Under the emerging paradigm of \textit{data perspectivism}, disagreement in annotation is treated as a meaningful signal rather than noise \cite{Cabitza_Campagner_Basile_2023}. Prior studies have explored various multi-annotator learning strategies, including incorporating annotator statistics and learning personal latent vectors \cite{kanclerz-etal-2022-ground}, learning multi-task and multi-label frameworks \cite{davani-etal-2022-dealing}, and annotator-aware representations \cite{mokhberian-etal-2024-capturing}. These approaches consistently show that modelling individual annotation patterns can achieve comparable and sometimes better performance to majority-vote baselines while better capturing uncertainty and variability in human judgement. Additionally, methods that leverage annotator metadata demonstrate that improvements often arise from learning annotator-specific behaviour rather than shared demographic patterns \cite{orlikowski-etal-2025-beyond}. Within MFT NLP, prior work has explored perspectivist approaches from different angles. \citet{golazizian-etal-2024-cost} explored cost-efficient approaches combine multi-task learning with few-shot annotator adaptation to incorporate new annotator perspectives while reducing annotation cost. In parallel, \citet{alvarez-nogales-araque-2024-moral} take an early step by training separate classifiers on subsets of annotators and combining their outputs via prompt-based ensemble; however, annotation sparsity and limited per-annotator data lead to substantial variability in annotator-specific predictions.

Collectively, this line of work highlights the importance of preserving perspectives in subjective tasks and motivates approaches, such as ours, that directly model individual annotators in MFT NLP and reveal insights into the impact of annotators and the subjective nature of moral foundations.  

\section{Data}
In this work we use the \textbf{Moral Foundations Twitter Corpus} (MFTC) \cite{MFTC}, comprising 35,108 tweets collected from Twitter (now X), across seven topics identified by hashtags. Twenty-three annotators were trained to manually assign labels for moral foundations and their polarity expressed in each separate tweets. An additional ``non-moral" label is included for annotators to flag tweets with no presence of any moral foundations. Every tweet received between 3 to 8 annotations, the majority receiving 3 or 4.

\begin{table}[h!]
    \centering
    \begin{tabular}{r|rr|r}
         Foundation & Moral & Absent & Proportion \\
         \hline
         Authority & 18473 & 109945 & 14.39 \\
         Care & 26141 & 102277 & 20.36 \\
         Fairness & 24635 & 103783 & 19.18 \\
         Loyalty & 17437 & 110981 & 13.58 \\
         Purity & 11982 & 116436 & 9.33 \\
    \end{tabular}
    \caption{Annotation distributions for all five foundations, with the proportion of annotations for moral classes (\textit{virtue} and \textit{vice}), showing class imbalance between the moral and absent classes.}
    \label{label dist}
\end{table}

Table \ref{label dist} shows the label distribution for all five foundations, where each foundation is considered for the entire dataset. There exists a substantial class imbalance between the moral classes and the absent class. 

\begin{table}[h]
    \centering
    \setlength{\tabcolsep}{3pt}
    \begin{tabular}{cccccc}
        \#Annotator & Mean & Median & Min & Max & s.d \\
        \hline
        23 & 5585 & 4588 & 560 & 19556 & 4747 \\
    \end{tabular}
    \caption{Summary statistics of the number of tweets each annotator has annotated, showing the sparsity of annotation structure.}
    \label{annotation table}
\end{table}

\begin{figure}[h]
    \centering
    \includegraphics[width=0.95\linewidth]{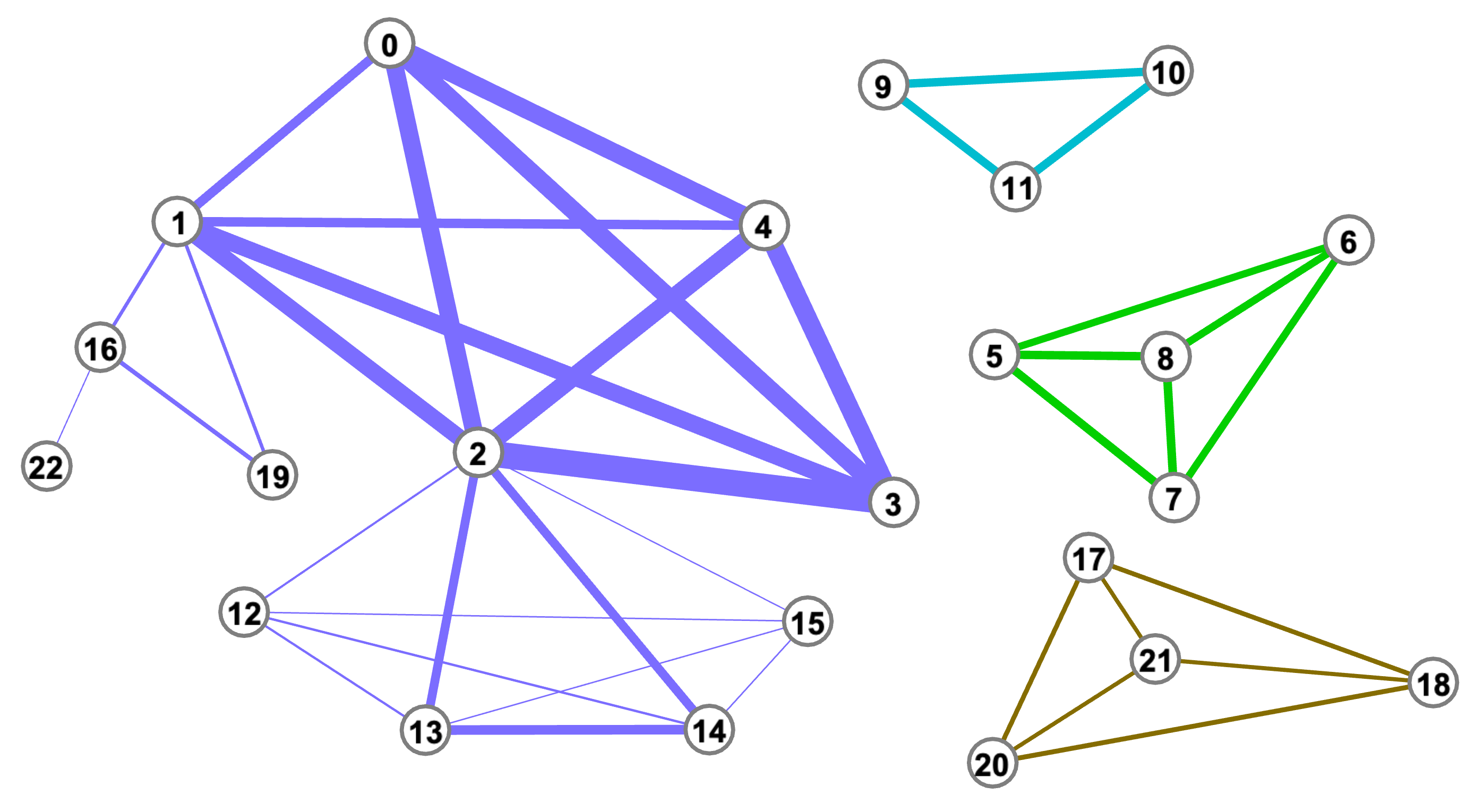}
    \caption{Co-annotation network between annotators. Edges represent the number of tweets co-annotated, and edges with weight less than 500 are filtered out. There exists several clusters of annotators which allows us to compare within groups of annotators that are labelled similar items. The disconnectedness shows that information is not shared across all annotators. }
    \label{coannotation}
\end{figure}

While the label distribution provides an overview of class prevalence, it does not capture how annotations are distributed across annotators. The variability in contributing annotations adds more complexity to the problem and we first show that with summary statistics in Table \ref{annotation table}. On average, each annotator labelled about one-seventh of the entire tweet collection. The large standard deviation highlights the variability of numbers of tweets each annotator labelled---a small number of annotators covered a majority of the dataset. This imbalance leads to sparse co-annotation, as visualised in Figure \ref{coannotation}. Edges in the graph have weights that represent the number of tweets two annotators both labelled and edges with weight less than 500 are removed to identify clusters where annotators within one cluster have been presented similar twitter content. The disconnectedness of the co-annotation network indicates that subsets of annotators never label any common items. As a result, information does not propagate across all annotators, preventing us from identifying global latent variables that helps explain annotator labelling ability and item labelling difficulty. This motivates the consideration of modelling approaches that are not restricted by disconnectedness and can make use of such locally structured annotation data, which we describe in the following section. 

Given the large amount of missing values in the annotation structure as shown above, we report inter-annotator agreement using Krippendorff's $\alpha$ \cite{krippendorff2018content} for each foundation in Table \ref{tab:agreement}. An $\alpha$ value of 0 indicates agreement at the level of chance, while a value of 1 indicates perfect consensus. We observe low to moderate agreement across all foundations, with \textit{Authority}, \textit{Loyalty} and \textit{Purity} exhibiting noticeably lower agreement than the others, suggesting a higher degree of subjectivity in their labelling. Overall, these highlight the need to explicitly capture diverse perspectives, as simple aggregation methods may be inadequate under such low levels of agreement. 

\begin{table}[h]
    \centering
    \setlength{\tabcolsep}{3pt}
    \begin{tabular}{ccccc}
         Authority & Care & Fairness & Loyalty & Purity \\
         \hline
          0.263 & 0.349 & 0.401 & 0.301 & 0.254
    \end{tabular}
    \caption{Krippendorff's $\alpha$ computed across 23 annotators per foundation, indicating low to moderate inter-annotator agreement and reflecting the subjective nature of moral-value labelling.}
    \label{tab:agreement}
\end{table}

Some prior work has disregarded the polarity of a foundation by merging virtue and vice labels or treated virtue and vice dimensions as two individual foundations. These approaches allow researchers to simplify the problem and avoid additional ``noise". However, we preserve the virtue and vice labels, as we aim to study the subjectivity of moral judgement not only identifying whether a foundation is expressed, but also in how annotators differ in assigning opposing moral valences.

\section{Methods}

\begin{figure*}[!t]
    \centering
    \begin{subfigure}[t]{0.48\textwidth}
        \centering
        \includegraphics[width=\linewidth,
        trim=0cm 12cm 10cm 0cm,
        clip]{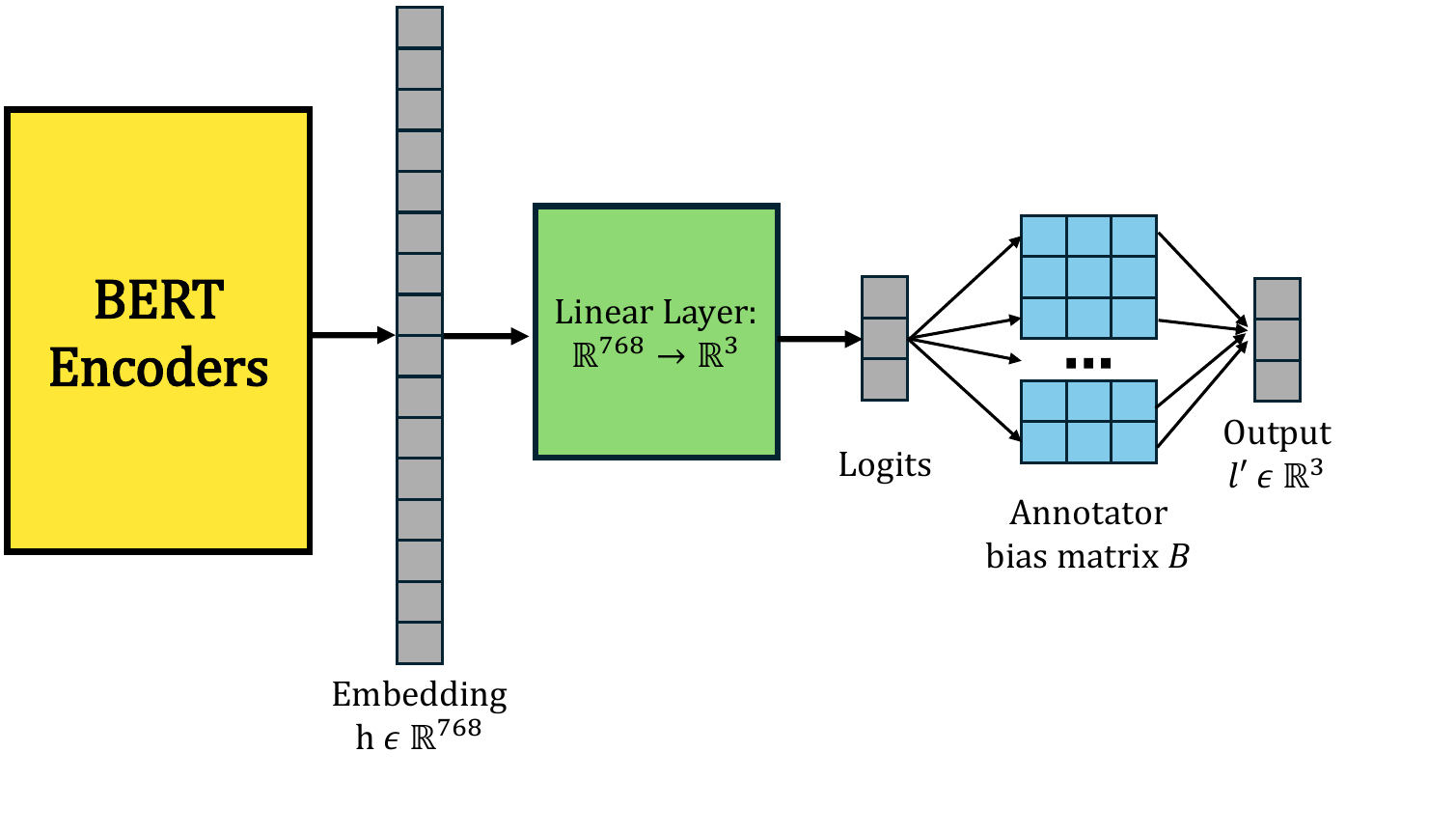}
        \caption{Bias-only variant that adjusts the final output probability based on the annotators' biases towards each class, yielding an interpretable bias matrix $B$.}
        \label{Bo-AL variant}
    \end{subfigure}\hfill
    \begin{subfigure}[t]{0.5\textwidth}
        \centering
        \includegraphics[width=\linewidth,
        trim=0cm 11.5cm 8cm 0cm,
        clip]{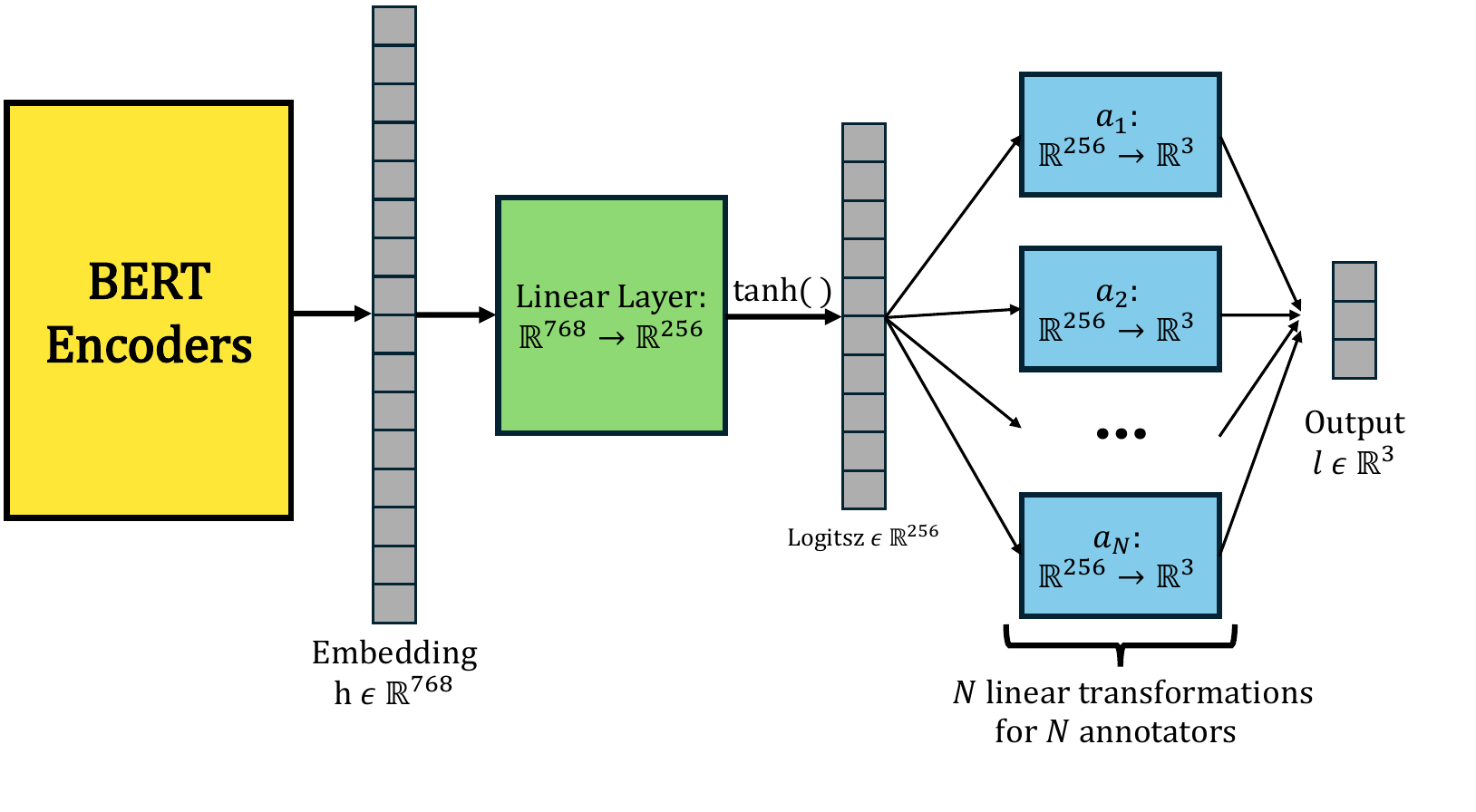}
        \caption{Linear transformation variant that expresses annotators' labelling patterns with linear layers, yielding better representation power of the individual perspectives.}
        \label{LT-AL variant}
    \end{subfigure}\hfill
    \caption{BERT finetuning with \textbf{Annotator Layer} that models annotators' individual labelling pattern and features.}
    \label{model sturcture}
\end{figure*}

\subsection{Problem Setup}
Our goal is to predict the moral foundation expressed in  text while accounting for individual perspectives of the human annotators. Each text instance $x_i$ is annotated by annotator $a_j$ in a group of $N$ annotators, producing a set of labels for all present moral values. For each foundation $k \in \{1, ..., 5\}$, we extract the label $y_{ij}^{(k)}$ and consider the tuple $(x_i, a_j, y_{ij}^{(k)})$ as a single observation. This way we keep annotators' individual labels rather than aggregating annotations into a single ``ground truth" label. We simplify the multi-label classification problem into single-label, multi-class classification tasks by considering each foundation separately. We train a separate classifier $f_k(x_i, a_j; \theta_k)$ for each of the five moral foundations (\textit{Authority, Care, Fairness, Loyalty, Purity}) to predict $y_{ij}^{(k)} \in \{1, 2, 3\}$, representing labels \textit{virtue, vice} and \textit{absent}.   

\subsection{Model Overview}
Our model extends a finetuned BERT classifier by incorporating neural network structures that learn annotator-specific features. This design is inspired by the \textit{Crowd Layer} framework, which applies neural network designs that directly learn from crowdsourced labels from multiple annotators \cite{crowdlayer}. Adapting this framework to our setting, we introduce the \textbf{Annotator Layer} that adjusts the shared text features according to the annotators, allowing the model to capture systematic difference in annotation patterns and annotators' individual perspectives. The model has 3 parts (Figure \ref{model sturcture}). Firstly, the pretrained language model BERT takes texts as inputs and outputs contextual embeddings.  We use \texttt{BERT-base-uncased} to encode text $x_i$, producing an embedding $h_i \in \mathbb{R}^{768}$ from the final hidden layer \texttt{[CLS]} token. The BERT model is finetuned with the following layers during the training process. The second component is a projection layer that maps the 768-dimensional BERT hidden representation to a lower-dimensional feature vector. The output dimensionality is determined by the following Annotator Layer variants.

\noindent \textbf{Bias-only} \hspace{0.5em} The objective of this variant is to provide interpretable annotator features learned from data. The projection layer has an output dimension of 3 that corresponds to the number of classes. It applies a linear transformation on the embedding $h_i$ to get the base logits $l_i$: 
$l_i = Wh_i + b.$

 For the Annotator Layer, we use an $N \times c$ matrix for $N$ annotators where each row of the matrix corresponds to the biases of an annotator towards each class. For the base logits $l_i$ with annotator id $a_j$, we adjust logits according to the annotator by computing $l_{ij} ' = l_i + B_j^T$,
 where $B \in \mathbb{R}^{N \times c}$ is the annotator bias matrix and $B_j^T$ denotes the transpose of the $j$-th row, representing annotator $a_j$'s bias across all classes. 

\noindent \textbf{Linear Transformation} \hspace{0.5em} The objective of this variant is to provide greater predicting power as we use much more parameters to represent the annotators. The projection layer has a tunable output dimension which we choose to use 256 as an intermediate value between 768 and 3. We apply a \textit{tanh} activation function to the output.
 For the Annotator Layer, each annotator $a_j$ is a linear transformation with a $3 \times 256$ weight matrix and a $3 \times 1$ bias vector. We compute the adjusted logits by 
$l_{ij} ' = W_{a_j}z_i + b_{a_j}$.

\noindent Both variants produce a 3 dimensional vector $l_{ij}'$, we then apply a \textit{softmax} function to yield the predicted probability distribution:
\[p_{ij} = \text{softmax}(l_{ij}'), \quad 
  \hat{y}_{ij}^{(k)} =  \arg\max_{c} p_{ij} ^{(c)},
\]
and the predicted class $\hat{y}_{ij}^{(k)}$ for text $x_i$ and annotator $a_j$ is the class with the greatest probability. 

\subsection{Training Objective}

We train the model with the cross-entropy loss and include two regularisation terms:
\begin{enumerate}
    \item \textbf{L2 Norm (Weight Decay):} standard L2 regularisation is applied to the model parameters to prevent overfitting.
    \item \textbf{Centred Bias Penalty:} For the bias-only variant of the Annotator Layer, we add a centred penalty on the bias matrix $B$:
    \[\mathcal{R}_{\text{centre}} = \left\| \frac{1}{N} \sum_{j=1}^{N} B_j \right\|_2^2.\]
    This is to ensure that the average annotator has zero bias towards each of the classes.
\end{enumerate}

In summary, the model combines a shared representation of each text with annotator-specific adjustments. BERT encodes the input into a contextual embedding, which is then mapped to base logits via a linear classification layer. The Annotator Layer modifies these logits according to the learned parameters of the annotator who supplied the label, modelling their individual perspectives. Via backpropagation, updates to the BERT parameters incorporate annotator information, enabling the encoders to refine text representations according to annotators’ various moral perspectives.

\section{Experiments}

Here we demonstrate two main aspects of our approach: (1) its effectiveness in predicting individual annotations, and (2) the interpretability of the learned annotator features. 

\begin{figure*}[!t]
    \centering
    \includegraphics[width=0.95\textwidth]{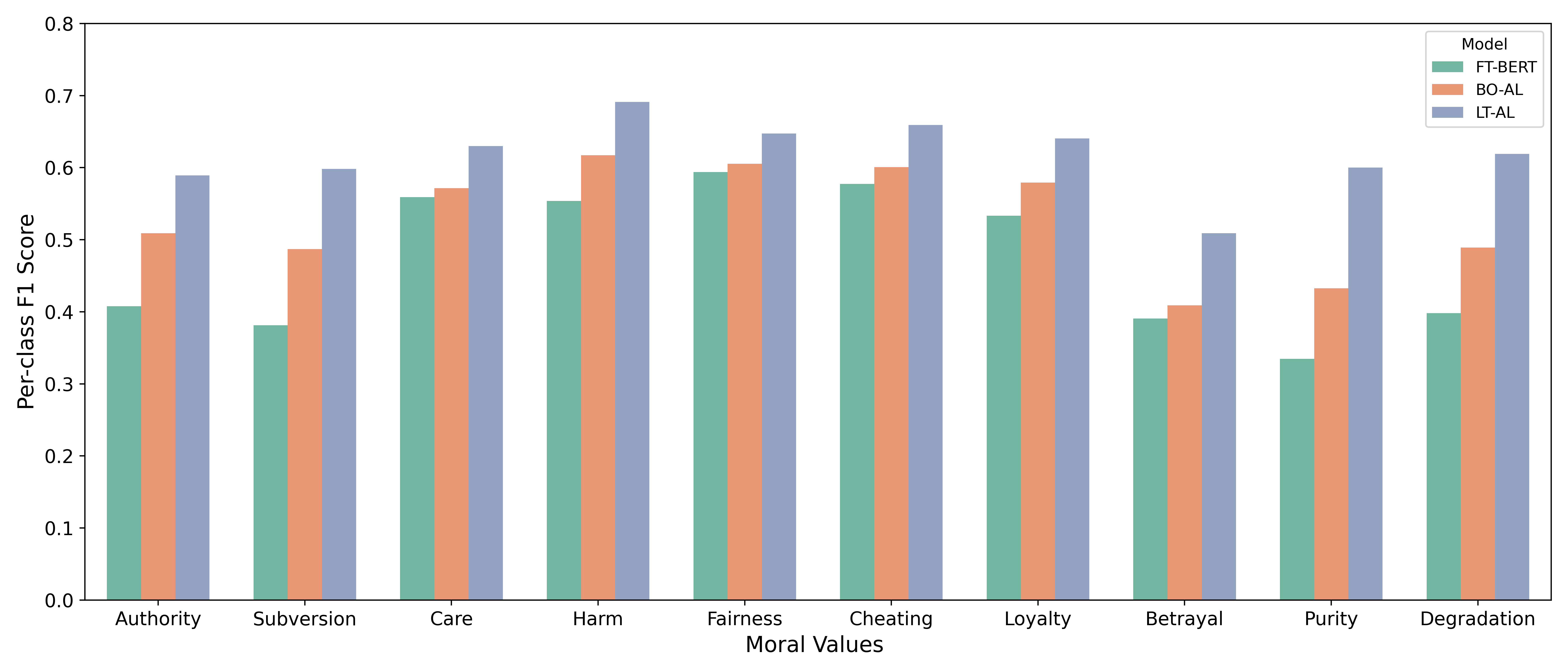}
    \caption{Average per-class F1 scores (five runs on different splits) for 10 moral values across five foundations (without class \textit{absent}), comparing three models: FT-BERT, BO-AL and LT-AL, evaluated on raw annotations. Both BO-AL and LT-AL models outperforms the baseline FT-BERT model.}
    \label{textann_performance}
\end{figure*}

We begin by preprocessing the texts following steps outlined in Appendix \ref{app:preprocessing}. We then separate the data into subsets that correspond to each of the five foundations, where we keep the cleaned texts, annotator ids and annotations. For each foundation, we create five folds using a \texttt{StratifiedGroupKFold} from scikit-learn \cite{scikit-learn}, which maintains the overall label distribution across folds and prevents data leakage by ensuring that all annotations belonging to the same tweet are grouped together in either the train or test set. Models are built and evaluated using the 5 non-overlapping splits and any metrics reported is an average score calculated across the five splits.

We compare the two variants of Annotator Layer (the Bias-only Annotator Layer (\textbf{BO-AL}) and Linear Transformation Annotator Layer (\textbf{LT-AL})) with a baseline of finetuned BERT without an Annotator Layer (\textbf{FT-BERT}). The latter adds a linear classifier that maps embeddings from BERT into a 3-class probability distribution and the parameters are finetuned. Finetuned BERT is currently regarded as the state-of-the-art approach in moral value classification. We deploy the same pretrained \texttt{BERT-base-uncased} and apply an identical training process wherever possible, allowing our experiments to also function as an ablation study. Training details and values of hyperparameters are recorded in Appendix \ref{app:training_details}. We report classification performance using F1 scores to better reflect performance (than classification accuracy) under the dataset's imbalanced label distribution.

\section{Results}
By modelling annotator-level features, Annotator Layers yield clear prediction improvements for individual annotations compared to the baseline.

\begin{table}[h]
    \centering
    \begin{tabular}{r|c|c|c}
            & FT-BERT & BO-AL & LT-AL \\
    \hline
    Authority & 57.3 & 64.6 & \textbf{71.3} \\
    Care      & 67.4 & 70.2 & \textbf{75.1} \\
    Fairness  & 69.7 & 71.1 & \textbf{74.9} \\
    Loyalty   & 62.1 & 64.3 & \textbf{70.0} \\
    Purity    & 56.3 & 62.7 & \textbf{72.9} \\
    \hline
    Overall   & 62.6 & 66.8 & \textbf{72.8} \\
    \end{tabular}
    \caption{Macro F1 scores for each foundation across three models: FT-BERT, BO-AL, and LT-AL. The addition of Annotator Layer improves classification performance, where the linear transformation variant yields the most improvement of 10.2\% in F1 score, averaged across all foundations.}
    \label{macro f1 table}
\end{table}

\begin{table*}[h]
    \centering
    \begin{tabular}{c|c|c|c|c|c|c|c|c|c}
            & \multicolumn{3}{c|}{Virtue} & \multicolumn{3}{c|}{Vice} & \multicolumn{3}{c}{Absent} \\
    \cline{2-10}
            & FT-BERT & BO-AL & LT-AL & FT-BERT & BO-AL & LT-AL & FT-BERT & BO-AL & LT-AL \\
    \hline
    A & 40.8 & 50.9 & 58.9 & 38.1 & 48.7 & 59.8 & 93.0 & 94.1 & 95.0 \\
    C      & 55.9 & 57.1 & 63.0 & 55.4 & 61.7 & 69.1 & 90.8 & 91.8 & 93.1 \\
    F  & 59.4 & 60.6 & 64.7 & 57.7 & 60.1 & 65.9 & 92.1 & 92.7 & 93.9 \\
    L   & 53.3 & 57.9 & 64.1 & 39.0 & 40.9 & 50.9 & 93.8 & 94.2 & 94.9 \\
    P    & 33.5 & 43.2 & 60.0 & 39.8 & 48.9 & 61.9 & 95.5 & 96.0 & 96.8 \\
    \end{tabular}
    \caption{Per-class F1 scores (Virtue, Vice, Absent) for each foundation (abbreviated by their initial letters) across three models: FT-BERT, BO-AL, and LT-AL. The greatest classification improvements occurs in the moral classes compared to the Absent class.}
    \label{perclass-f1 table}
\end{table*}

Figure \ref{textann_performance} shows that as we increase the complexity of neural network structures that represent the annotators (from none for FT-BERT, to linear layers for LT-AL), the performance improves across all foundations. With more model parameters, the LT-AL model has greater representational power for modelling individual annotators, leading to a largely improved classification performance over the baseline when predicting individual annotations. Even the BO-AL model yields a clear performance gain, despite adding only a small bias matrix (69 parameters). We observe particularly large improvement in F1 scores for \textit{Authority}, \textit{Loyalty} and \textit{Purity}. Table \ref{macro f1 table} further validates these results, showing improvements in macro F1 scores across all foundations when adding Annotator Layers.

Table \ref{perclass-f1 table} shows the greatest classification performance gains occur in moral classes (\textit{virtue} and \textit{vice}). For instance, the greatest improvement over the baseline model is the classification of \textit{purity} (virtue aspect of foundation \textit{Purity}) with the LT-AL model, yielding an increase of 0.265 in F1 score. All three models perform comparably when it comes to classifying the \textit{absent} class (a tweet that is not expressing a certain moral value). However, we still see an improvement for the \textit{absent} class with the addition of Annotator Layers, even when the baseline is already sufficiently strong.

\section{Discussion}

\subsection{Interpretability of Annotator Layer}

\begin{figure}[h]
    \centering
    \includegraphics[width=0.95\linewidth, trim={0 1cm 0 1.5cm}]{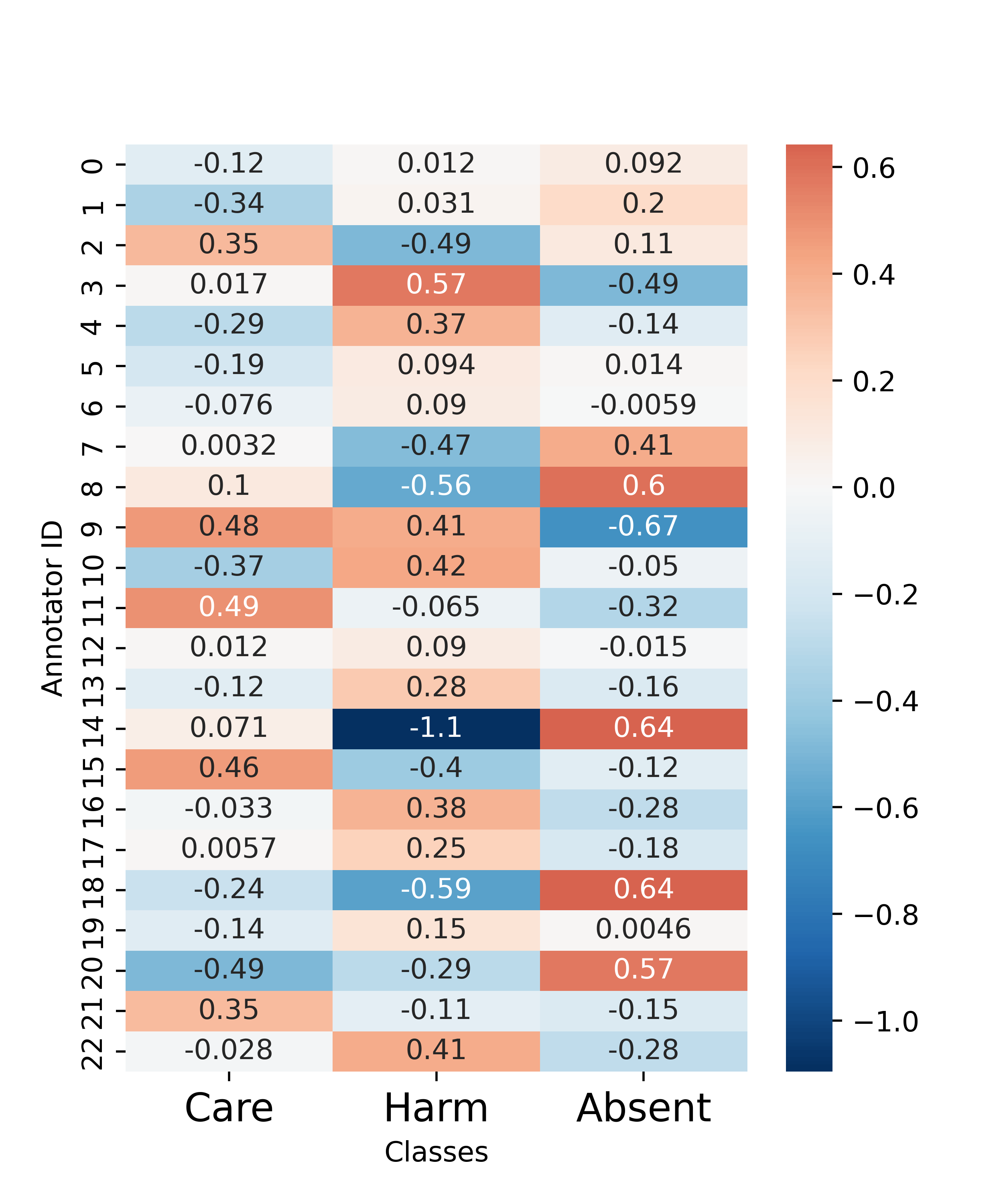}
    \caption{Bias matrix for the Care foundation extracted from the bias-only Annotator Layer. Positive values (see colourbar) indicate positive bias. This matrix gives information regarding how annotators give labels with different tendencies for the same foundation.}
    \label{care bias}
\end{figure}

To illustrate the interpretability of the bias-only Annotator Layer, we extract the bias matrix from the trained models and analyse annotators' biases towards each class. Figure \ref{care bias} visualises annotators' biases per foundation level. Positive values indicate biases towards a class whereas negative values indicate biases against a class. Several bias patterns are observed in the bias matrix; we use the \textit{Care} foundation as an example. Annotator 2 shows a moderate bias toward the virtue aspect \textit{Care} and, correspondingly, a bias against the vice aspect \textit{Harm}. However, this complementary behaviour between the two polarities does not always hold. When an annotator possesses a tendency towards or against one moral class, the complementary class may instead be \textit{absent}. We observe this pattern in several annotators (e.g. Annotator 3 and 7). In some cases, annotator exhibit biases towards or against both moral classes, with the \textit{absent} class acting as the complement.

\begin{figure}[h]
    \centering
    \includegraphics[width=0.49\linewidth]{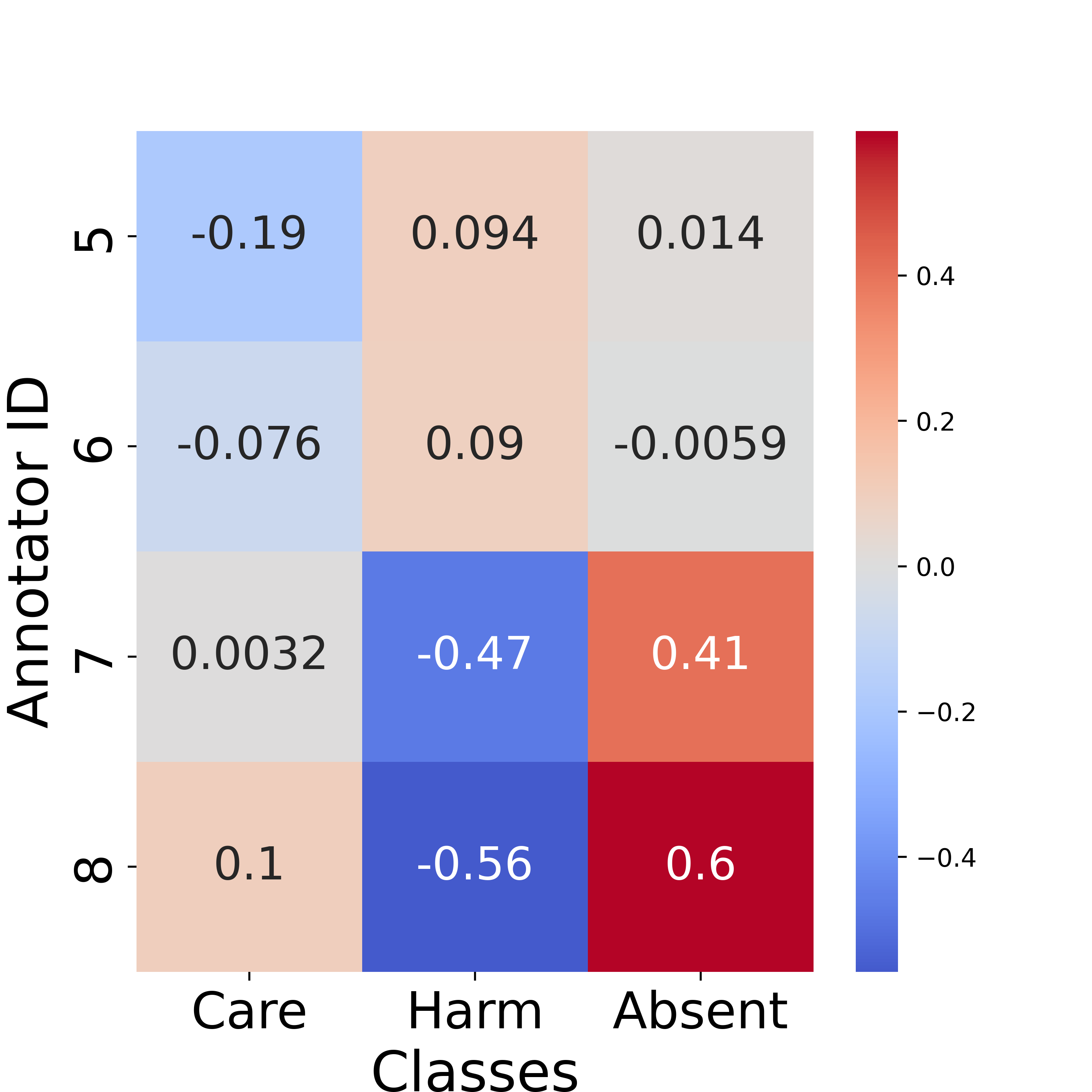}
    \hfill
    \includegraphics[width=0.49\linewidth]{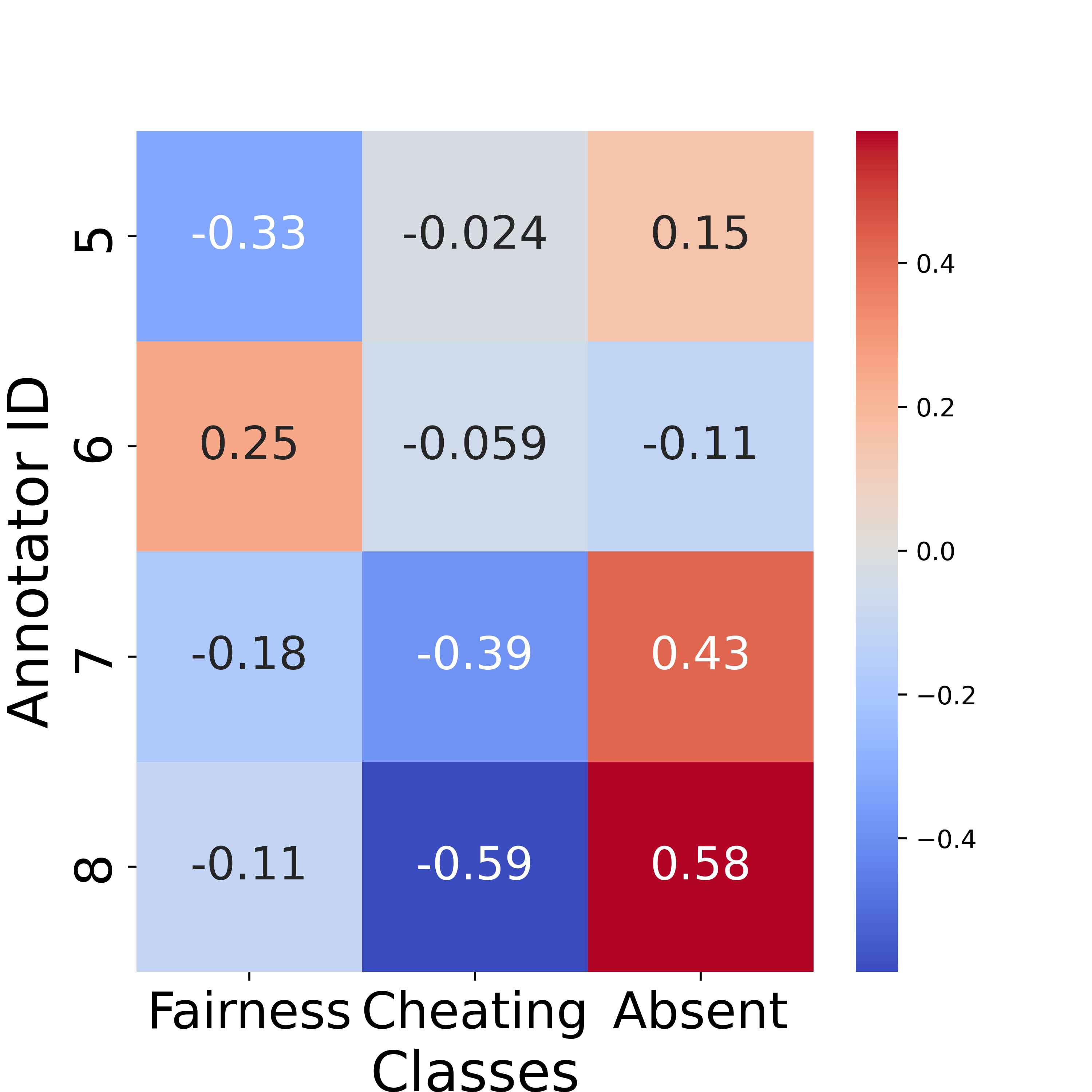}
    \caption{Bias Heatmaps of Foundations Care (Left) and Fairness (Right), for Annotators 5, 6, 7 and 8, showing annotators with similar bias patterns in one foundation can have opposite bias patterns in another. }
    \label{5678 bias}
\end{figure}

We also show bias weights for a subset of annotators for the \textit{Care} and \textit{Fairness} Foundations. Annotators 5, 6, 7 and 8 belong to the same co-annotation cluster (Figure \ref{coannotation}) meaning they were presented similar tweet contents. Hence, we pick this group to show how the labelling patterns share similarity and difference across foundations. Figure \ref{5678 bias} shows the biases over two foundations for the group of selected annotators. We observe that Annotators 7 and 8 show a consistent tendency to favour the \textit{absent} class, with only minor differences in their biases toward the \textit{virtue} classes across the two foundations. In contrast, Annotators 5 and 6 both have different tendencies between the two foundations. Both show similar patterns for \textit{Care}, favouring the \textit{vice} class and labelling against the \textit{virtue} class. However, for \textit{Fairness}, they exhibit opposite patterns. Annotator 5 favours the \textit{absent} label and gives fewer \textit{virtue} labels, whereas Annotator 6 shows the reversed behaviour. These observation suggests that, while some bias patterns are consistent, they should be analysed separately for each foundation rather than assuming a universal annotator bias.

By summarising these observations, we may identify groups of annotators who share similar perspectives when interpreting moral values for each foundation. These groups exhibit consistent patterns in moral judgement, such as virtue/vice-oriented annotators, annotators who frequently give moral labels, and those who give moral labels more cautiously, leading to dominating non-moral (absent) annotations. Such patterns suggest meaningful ``annotator types", revealing insights into the diversity of moral judgement and providing potential categorisation for all individuals, not just annotators. This provides possible modelling approaches that learn group behaviours instead of modelling individual annotators, such as mixture-of-experts models where experts represents groups of people with similar perspectives. One can also study correlations between individual political ideology, cultural background and other demographic factors with the ``types" that we identify, gaining insights into the development of diverse perspectives. 

To examine whether these bias patterns capture information beyond simple annotator-level label preferences, we construct a non-learned empirical baseline using relative label preferences $P(c|a)/P(c)$, where each annotator's labelling distribution is normalised by the overall class distribution to account for class imbalance. Replacing the learned bias matrix with this empirical preference yields comparable performance, suggesting that both approaches shift the predicted probability distribution in similar directions for each annotator. However, we observe that while the direction of these shifts is broadly aligned, the magnitude of the learned biases differs from the empirical preference matrix. This indicates that the learned bias is not merely reproducing simple dataset statistics. The Jensen-Shannon divergence between the learned bias and empirical preference matrices averages $0.446 \pm 0.070$ across annotators, ranging from $0.329$ to $0.562$, indicating that the two representations are not closely aligned at the distribution level. Moreover, the text encoder is jointly finetuned with the bias matrix and may encode annotator related information, which is not isolated in this analysis. These results suggest that, despite similar empirical performance, the bias-only model still captures meaningful structure beyond label frequency statistics.

\subsection{Raw Annotations and Aggregated Labels}

Using the trained models with Annotator Layers, we apply a simple rule to obtain an aggregated label for each tweet. For each tweet, we activate all ``annotators" in the Annotator Layer, regardless of their IDs, and obtain 23 predictions, yielding 23 probability distributions over the 3 classes. We then average these distributions and select the class with the highest mean probability as the aggregated prediction. For this analysis, we train the baseline model FT-BERT directly on aggregated labels, mimicking the common practice used when fine-tuning BERT for moral value classification. 

\begin{figure}[h]
    \centering
    \includegraphics[width=\linewidth]{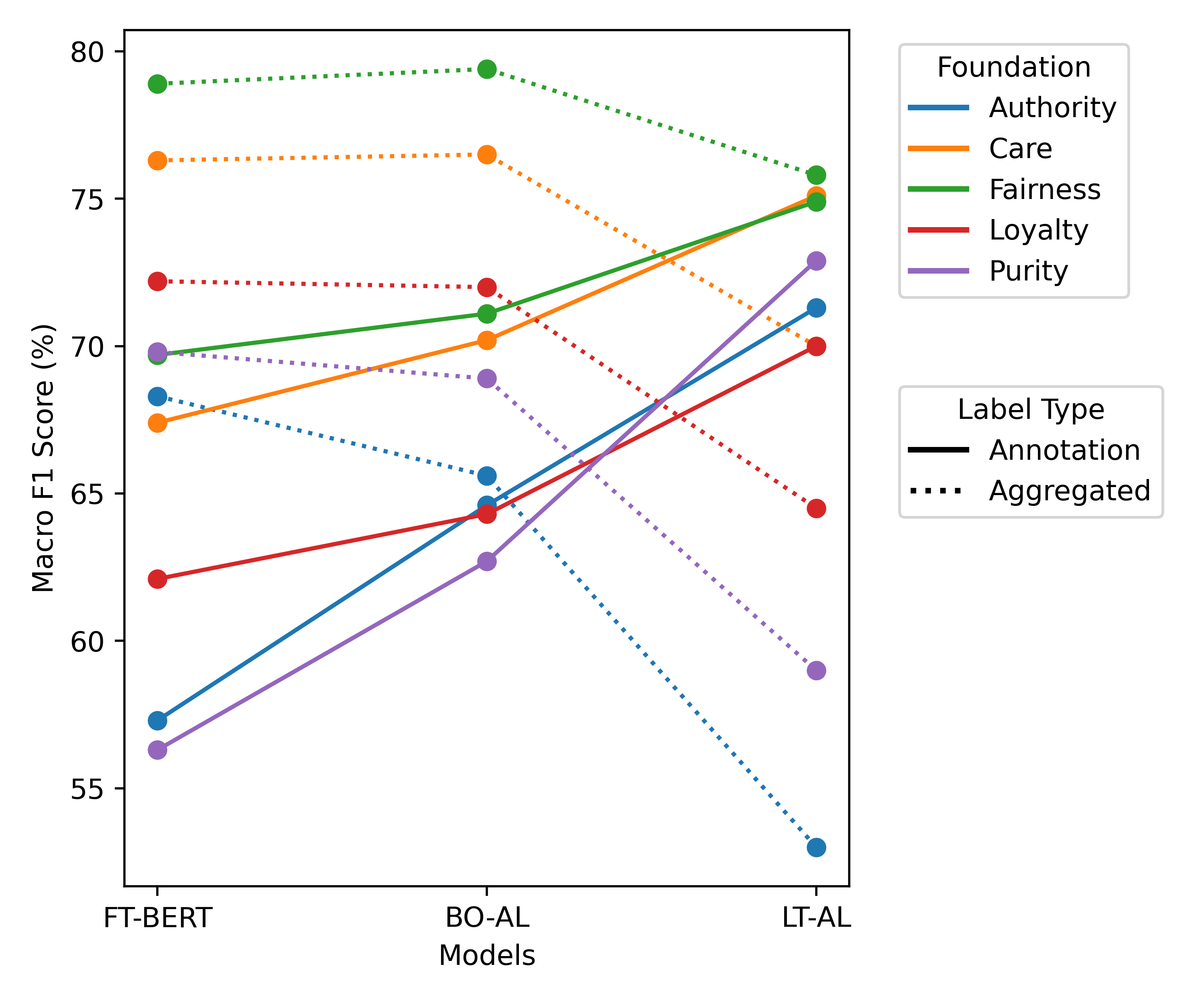}
    \caption{Trend of macro F1 scores as the modelling capacity to represent the annotators increases, compared for predicting raw annotations and aggregated labels. As the capacity to model individual perspectives increases (from FT-BERT to LT-AL), we see an increase in classification performance on raw annotations and a decrease in classification performance on aggregated labels. }
    \label{trend}
\end{figure}

We show the changes in macro F1 scores in Figure \ref{trend} as we move from standard finetuned BERT, to the two variants of the Annotator Layers. The horizontal axis can be interpreted as an increasing capacity to model annotator-specific features from left to right. We've already shown in the \textbf{Results} section that as we increase the capacity, the prediction accuracy of raw annotations also increases (solid lines in Figure \ref{trend}). When evaluating on aggregated labels, the FT-BERT and BO-AL models achieve comparable performance, whereas the LT-AL model shows a substantial decrease. This decrease when evaluating on aggregated labels is more obvious where we see the largest gain when evaluating on raw annotations. Interestingly, the transition from BO-AL to LT-AL yields the strongest improvement on raw annotations, but it also produces substantial decrease under aggregated-label evaluation. This is expected, the LT-AL models are designed to capture annotator-specific features and better learn how individuals interpret texts and assign labels. When activating all annotators and collapsing their outputs into a “consensus’’ label, additional noise is introduced due to the sparse annotation structure. In essence, we are asking the learned annotator representations to make predictions on tweets that the corresponding annotators never see, with potentially great domain differences. These observations demonstrate that a model that is capable of representing annotators' individual perspectives does not necessarily agree with the aggregated labels. This highlights an important limitation of training models on aggregated labels: a strong performance by such a model may hide substantial underlying variations in individual perspectives and may not reflect the true effectiveness of the model. 

\subsection{Ambiguity of Foundations}

\begin{table}[h]
    \centering
    \begin{tabular}{r|r|rrr}
    Foundation & Type & Virtue & Vice & Mean \\ 
    \hline
    Care & I & 7.1 & 13.7 & 10.4 \\
    Fairness & I & 5.3 & 8.2 & 6.8 \\ 
    Authority & B & 18.1 & 21.7 & \textbf{19.9}\\
    Loyalty & B & 10.8 & 11.9 & \textbf{11.4} \\
    Purity & B & 26.5 & 12.1 & \textbf{19.3} \\
    \end{tabular}
    \caption{Performance improvement in F1 scores (\%) between LT-AL and FT-BERT for the five moral foundations. The binding (B) foundations benefits more from modelling annotators' individual perspectives when compared to the individualising (I) foundations.}
    \label{gains}
\end{table} 

While the addition of Annotator Layers improves overall performance, the gains vary across foundations, suggesting that some moral foundations exhibit greater ambiguity and therefore benefit more from annotator-specific modelling. Greater improvements occur in \textit{Authority}, \textit{Loyalty}, and \textit{Purity}, compared to \textit{Care} and \textit{Fairness} (Table \ref{gains}). This pattern mirrors the distinction between individualising foundations (\textit{Care, Fairness}) and binding foundations (\textit{Authority, Loyalty, Purity}). Individualising foundations are generally considered more morally relevant and are endorsed across the political spectrum, whereas binding foundations tend to receive endorsement from a smaller portion of the population \cite{Graham2009-da}. We've shown that human annotators exhibit lower agreement on the binding foundations (Table \ref{tab:agreement}), as measured by Krippendorff's $\alpha$, indicating greater ambiguity. This pattern align with both the observed performance gains and the psychological distinctions between binding and individualising foundations.

\section{Conclusion}

In this work we introduced the Annotator Layer for moral classification of texts that captures annotator-specific moral perspectives and annotation patterns, extending on finetuning BERT models. Our experiments demonstrate improved classification performance of individual annotations in crowdsourced dataset, along with interpretable representations of annotators' bias patterns. It is shown that disagreement between annotators in such subjective tasks is a learnable feature instead of annotation noise. The results suggest that relying solely on aggregated labels can hide important information. We hope this work encourages future research to move beyond training a universal classifier that predicts a ``ground truth" and develop models that better reflect diversity of moral judgement and understand the subjectivity of moral classification of texts. 

\section{Limitations}

Our work has two primary limitations.

First, although the dataset publication notes that annotator metadata (e.g., demographic information, political ideology and moral values measured by MFQ) exists, this information was not available to us and is therefore not incorporated into the analysis. As a result, while the Annotator Layer learns annotator-specific features and identified potential differences of annotation patterns between groups of annotators, we cannot directly examine how these patterns relate to known characteristics. Studies in moral psychology have validated that these characteristics have a direct impact to human moral judgement. Hence, access to such data can help validate the learned representations and provide explanations to some of the observed patterns. 

Second, our approach does not provide a strong mechanism for producing high-quality aggregated predictions to moral values. We've demonstrated the bias-only variant's comparable classification performance to finetuned BERT on aggregated labels, but the linear transformation variant has shown a substantial decrease in performance. Many downstream applications ultimately requires a single, aggregated label for a text observation, yet our annotator-specific models requires annotator (human) information to provide accurate predictions which typically lacks in these tasks. The model learns fine-grained human-specific behaviours and does not generalise well for aggregated labels. Our naive approach to obtain consensual labels by activating all annotator corresponding neural network structures and averaging the prediction distributions introduces noise, especially given the sparse and uneven co-annotation structure. Developing principled aggregation methods that leverage annotator features is a vital future direction.

\paragraph{Acknowledgements}
This work was supported with supercomputing resources provided by the Phoenix HPC service at Adelaide University. 


\bibliography{ref}

\appendix

\section{Training Details}

\subsection{Text Preprocessing}
\label{app:preprocessing}
We clean the twitter texts by first removing URLs, non-alphanumeric characters, punctuations and retweet markers. All text is then lowercased, and user mentions are replaced with the token “@user”. Stopwords may optionally be removed, though we found this to have negligible effect on model performance.

\subsection{Training Process and Hyperparameters}
\label{app:training_details}
We implement and train all models using {Pytorch} \cite{Pytorch} {v2.7} and optimise the parameters using the {AdamW} optimiser \cite{AdamW}. The initial learning rate is 2e-5 for the BERT parameters and is 1e-4 for the parameters in the linear layer and Annotator Layer, with linear decay and no warm-up. The lower learning rate is used to update the parameters of BERT moderately, avoiding deterioration of BERT's ability of capturing semantic and contextual meaning with the embeddings. In training, we use a batch size of 8, and the maximum input text length is set to be 64 tokens as all texts in the dataset are short in length. We set the L2 regularisation coefficient to 0.01 and the centred bias penalty to 0.05. We train the models for 5 epochs and freeze the BERT parameters during the first epoch to allow the newly added layers to stabilise before full finetuning. All experiments are run using one Nvidia A100-SXM4-40GB GPU. 

\end{document}